\documentclass[conf]{new-aiaa}
\usepackage[utf8]{inputenc}

\usepackage{graphicx}
\usepackage{amsmath}
\usepackage[version=4]{mhchem}
\usepackage{siunitx}
\usepackage{longtable,tabularx}
\usepackage{subcaption}
\title{Application of Convolutional Neural Network\\to Predict Airfoil Lift Coefficient}

\author{Yao Zhang\footnote{Ph.D.Candidate, yzhang401@gatech.edu, Student Member AIAA}, WoongJe Sung\footnote{Research Engineer II, woongje.sung@asdl.gatech.edu, Member AIAA}, Dimitri Mavris\footnote{Regents Professor for Advanced Systems Analysis, AIAA Fellow}}
\affil{School of Aerospace Engineering, Georgia Institute of Technology, Atlanta, GA, 30332}

\begin{document}

\maketitle

\begin{abstract}
The adaptability of the convolutional neural network (CNN) technique is probed for aerodynamic meta-modeling task. The primary objective is to develop a suitable architecture for variable flow conditions and object geometry, in addition to identifying a sufficient data preparation process. Multiple CNN structures were trained to learn the lift coefficients of the airfoils with a variety of shapes in multiple flow Mach numbers, Reynolds numbers, and diverse angles of attack. This was conducted to illustrate the concept of the methodology. Multi-layered perceptron (MLP) solutions were also obtained and compared with the CNN results. The newly proposed meta-modeling concept has been found to be comparable with the MLP in learning capability; and more importantly, our CNN model exhibits a competitive prediction accuracy with minimal constraints in geometric representation.
\end{abstract}

\section{Nomenclature}

{\renewcommand\arraystretch{1.0}
\noindent\begin{longtable*}{@{}l @{\quad=\quad} l@{}}
$\alpha$, AoA & angle of attack \\
$\rho$ & raw pixel density \\
$\bar{\rho}$ & pixel density \\
$c$ & chord length \\
$C_l$ & sectional lift coefficient \\
CFD & computational fluid dynamics \\
CNN & convolutinal neural network \\
FC & fully-connected \\
$M_\infty$ & freestream Mach number \\
MLP & multi-layered perceptron \\
MSE & mean squared error \\
NACA & National Advisory Committee for Aeronautics \\
ReLU & rectified linear unit \\
tanh & tangent sigmoid function \\
trn & dataset for training \\
vld & dataset for validation
\end{longtable*}}

\section{Introduction}
\lettrine{T}{he} rise in computational power has undeniably transformed engineering analysis and design. Recently, the rapid advance in data science and machine learning techniques has opened new doors in understanding unsolved aerodynamics and fluid mechanics problems \cite{zhang2015machine,parish2016paradigm,muller1999application}. It also allows for more possibilities in terms of predictive and control capability \cite{gautier2015closed}. The traditional way of solving aerodynamics and fluid mechanics problems is a top-down approach that relies on physics modeling. In order to build a good model, a researcher needs to have a clear understanding of physics and theories. Most of these problems are high-dimensional, multi-scale, and nonlinear problems, which can be very difficult to find underlying physics. Some low-fidelity models are based on a first-order understanding. However, due to the complexity of the real-world problems, there will inevitably be non-negligible differences between the actual dynamics and the approximated models. Thus, the goal of meta-modeling, in this context, is to balance the modeling efforts and scope of the design space. This is a difficult task \cite{sung2012neural} that begs the question of whether there is a way to look at the problem from a different perspective. It is believed that with the help of data science and machine learning, there may be a way.

For the past decade, the potential of combining the meta-modeling of CFD tools with machine learning techniques has gained widespread attention. Data-driven surrogate models are becoming popular, due to the high volume of physical testing and the simulation data generated from the design, analysis and optimization processes. Different machine learning algorithms are being proposed to solve aerodynamics problems. In addition, deep learning techniques have been brought to many researchers' attention \cite{zhang2015machine,tracey2013application}. Most of the existing interdisciplinary work, in terms of the combination of meta-modeling of CFD and machine learning techniques, adopts the learning architecture that belong to the category of the `vanilla' multi-layered perceptron (MLP). MLP is a straightforward, fully-connected neural network architecture that is not specifically designed to exploit spatial and/or temporal correlation that are intrinsic in many real-world problems. For example, M\"{u}ller et al. applied MLP approaches to construct low-order models describing the near wall dynamics in turbulent flows \cite{muller1999application}. Rai and Madavan have shown the potential of using multi-layered neural networks with the 2D aerodynamic design of turbomachinery \cite{rai2001application,rai2000aerodynamic}. Their results demonstrate the advantage of the capability and efficiency to achieve design targets through the MLP architecture. The MLP learning technique does work, but there still is much room for improvement. This improvement may include faster training and reduction in the amount of simulation data required for prediction accuracy in a design space \cite{rai2001application}. In the MLP architecture, the learning capability can be increased by additional hidden layers and/or additional hidden units in each hidden layer. However, the trade-off space between the network size and the learning capability is highly quantized with stereotyped variability due to the underlying assumption of `fully-connected' network structure. In the other hand, the convolutional neural network (CNN) provides an attractive alternative to the MLP and rapidly replacing the standard MLP technique in many challenging machine-learning tasks such as image recognition \cite{lecun1998gradient,lecun2015deep}.
In this investigation, we propose a methodology applying the convolutional neural network (CNN) framework to the meta-modeling of fluid mechanics analysis tools. Previous work \cite{zuo2015convolutional,taylor2010convolutional} has illustrated that the CNN enables to learn invariant high-level features when the data have strong spatial and/or temporal correlations. This ability is crucial for many aerodynamic meta-modeling tasks as much as for other real-world learning tasks. Guo et al. applied CNN to approximate the steady state laminar flow \cite{guo2016convolutional}. This work has shown that CNNs enable a rapid estimation for the flow field, and the study focused on the visual replication of the velocity field around a group of categorical geometries. Yilmaz and German \cite{yilmaz2017convolutional} applied CNN architecture to map airfoil shapes to pressure distribution under the framework of classification problem using discretized pressure coefficient. This study learned and predicted pressure distribution on airfoils in zero-angle of attack only. These recent work has demonstrated the increasing attention to the CNN techniques in the fluid mechanics domain. This attention partly originates from the potential benefit of the CNN's flexibility in geometric representation and scalability for the larger problems beyond current capabilities. \emph{In this work, multiple CNN structures were trained to predict the lift coefficients of airfoils with a variety of shapes in multiple freestream Mach numbers, Reynolds numbers, and diverse angles of attack.}

\section{Methods}
\subsection{Convolutional Neural Network Architecture}
Since the seminal introduction of the LeNet-5 architecture \cite{lecun1998gradient}, the CNN has widely been used for challenging machine-learning tasks, especially in image recognition applications. It is one of the core computational structures enabling a significant share of modern deep learning techniques \cite{lecun2015deep}. The CNN can capture the invariance from complex and noisy real-world data in a scalable manner, due to a unique sub-structure of local receptive fields with weight-sharing and subsequent mapping through sub-sampling layers \cite{haykin2001neural}.

\begin{figure}[hbt!]
\centering
\includegraphics[width=.75\textwidth]{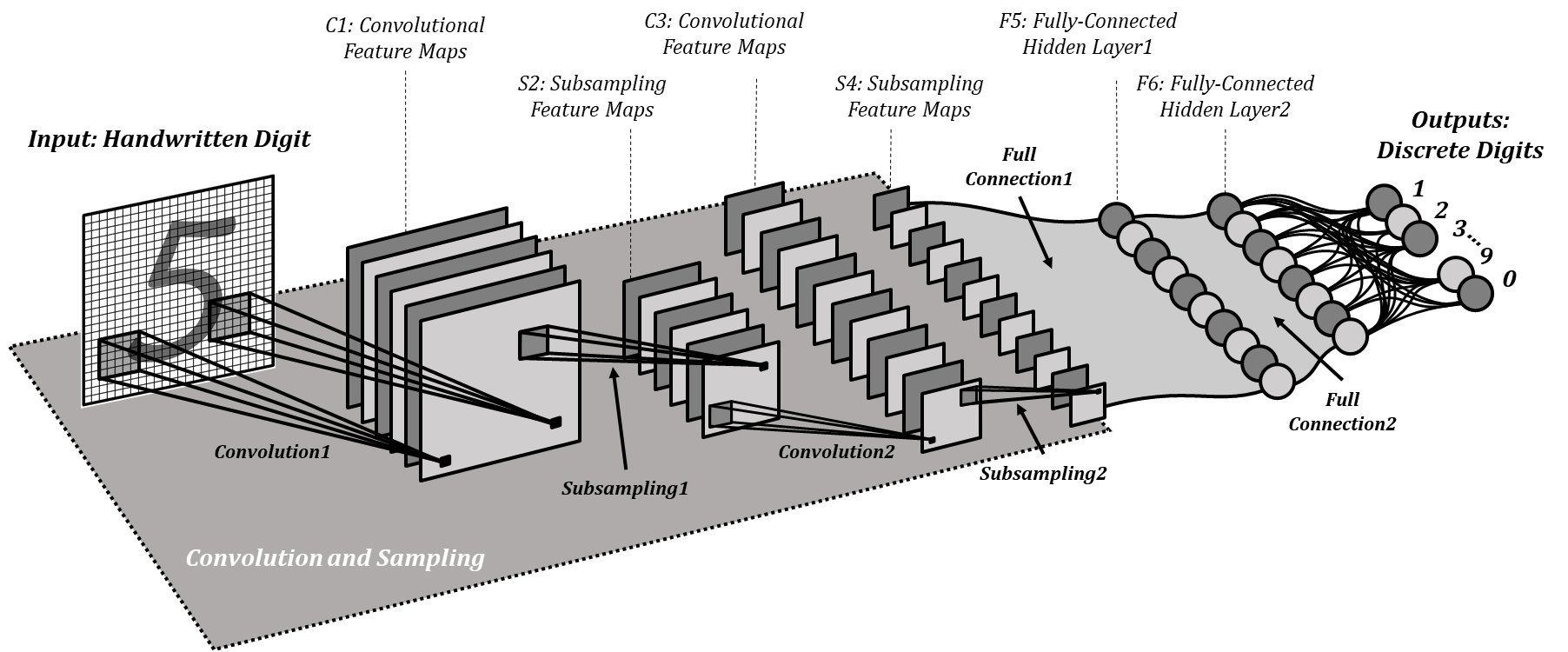}
\caption{LeNet-5\cite{lecun1998gradient}: CNN for handwritten digit recognition task}
\label{fig:lenet5}
\end{figure}

\begin{figure}[hbt!]
\centering
\includegraphics[width=.80\textwidth]{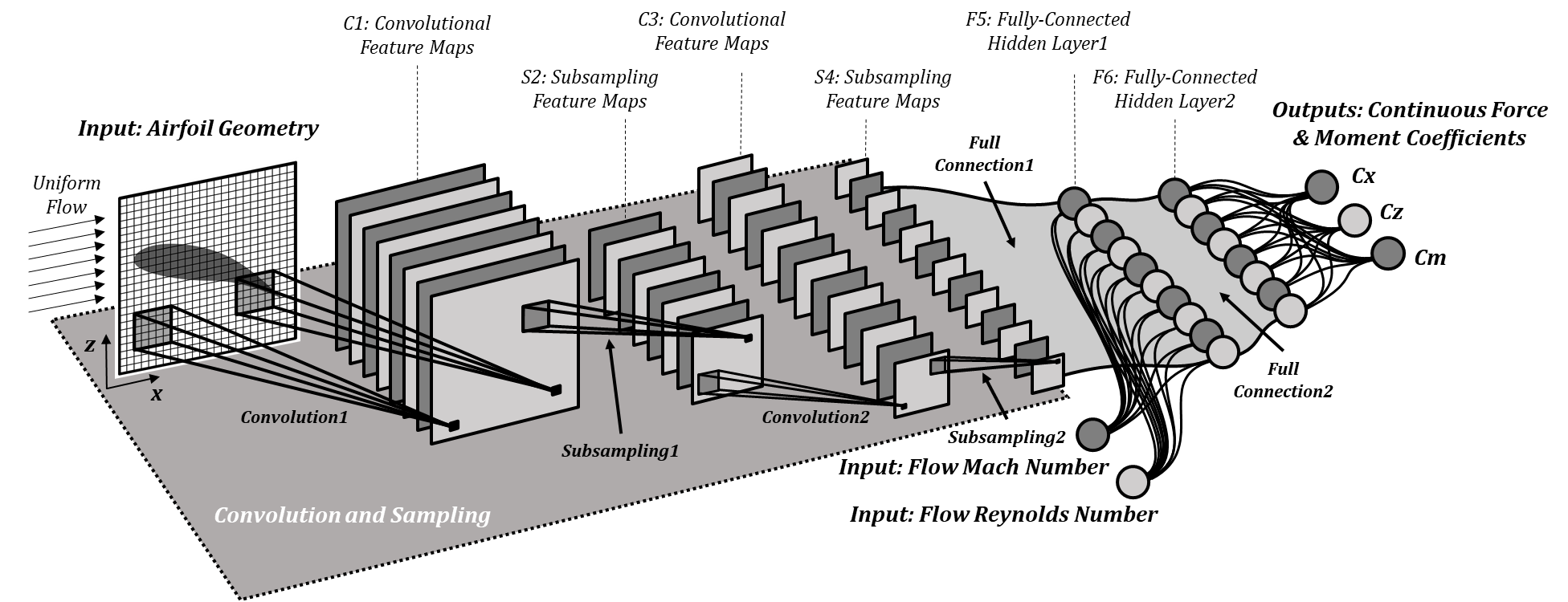}
\caption{Modified LeNet-5 for a notional aerodynamic meta-modeling task}
\label{fig:mlenet5}
\end{figure}

Fig. \ref{fig:lenet5} illustrates the simplified layout of the LeNet-5 applied to the handwritten digit recognition task \cite{lecun1998gradient}. Beyond this structure, aerodynamic meta-modeling requires various modifications. Besides object geometry, flow conditions need to be considered. In addition, the network output needs to be a continuous regression instead of a discrete classification. Fig. \ref{fig:mlenet5} is a conceptual CNN architecture reflecting those modifications on top of the LeNet-5 where, in contrast to the conventional aerodynamic meta-modeling, the object geometry is represented by an image-like array of pixels instead of a set of coordinates, shape functions, or pre-defined parameters.

\begin{figure}[h!]
\centering
\includegraphics[width=0.75\textwidth]{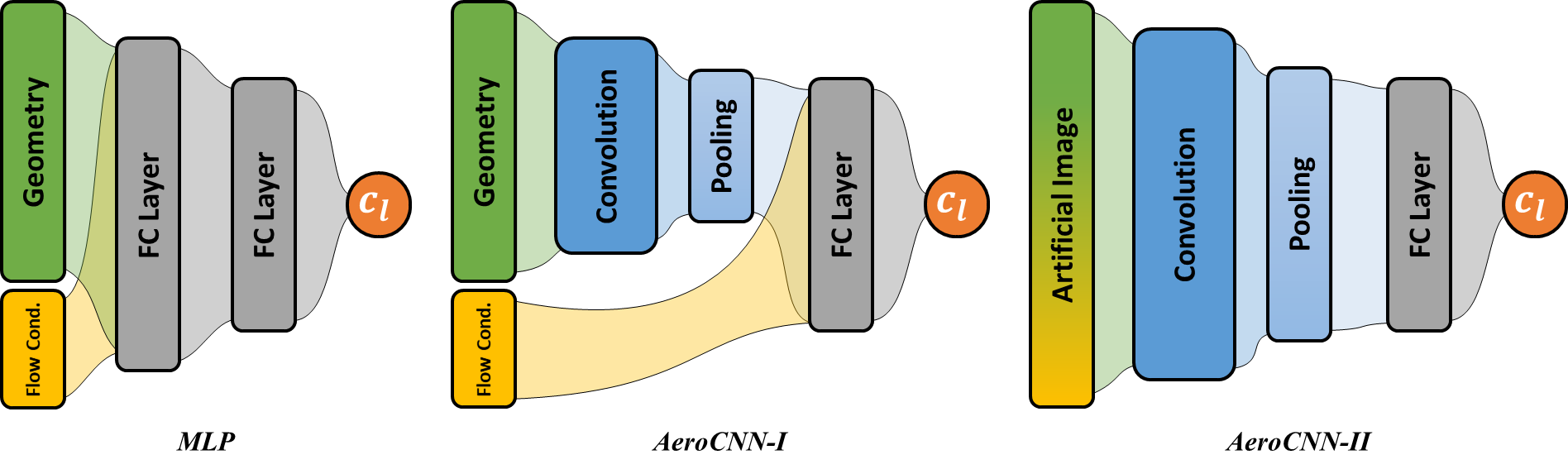}
\caption{Three network structures tested in this work}
\label{fig:3nets}
\end{figure}

\begin{figure}[h!]
\centering
\includegraphics[width=0.75\textwidth]{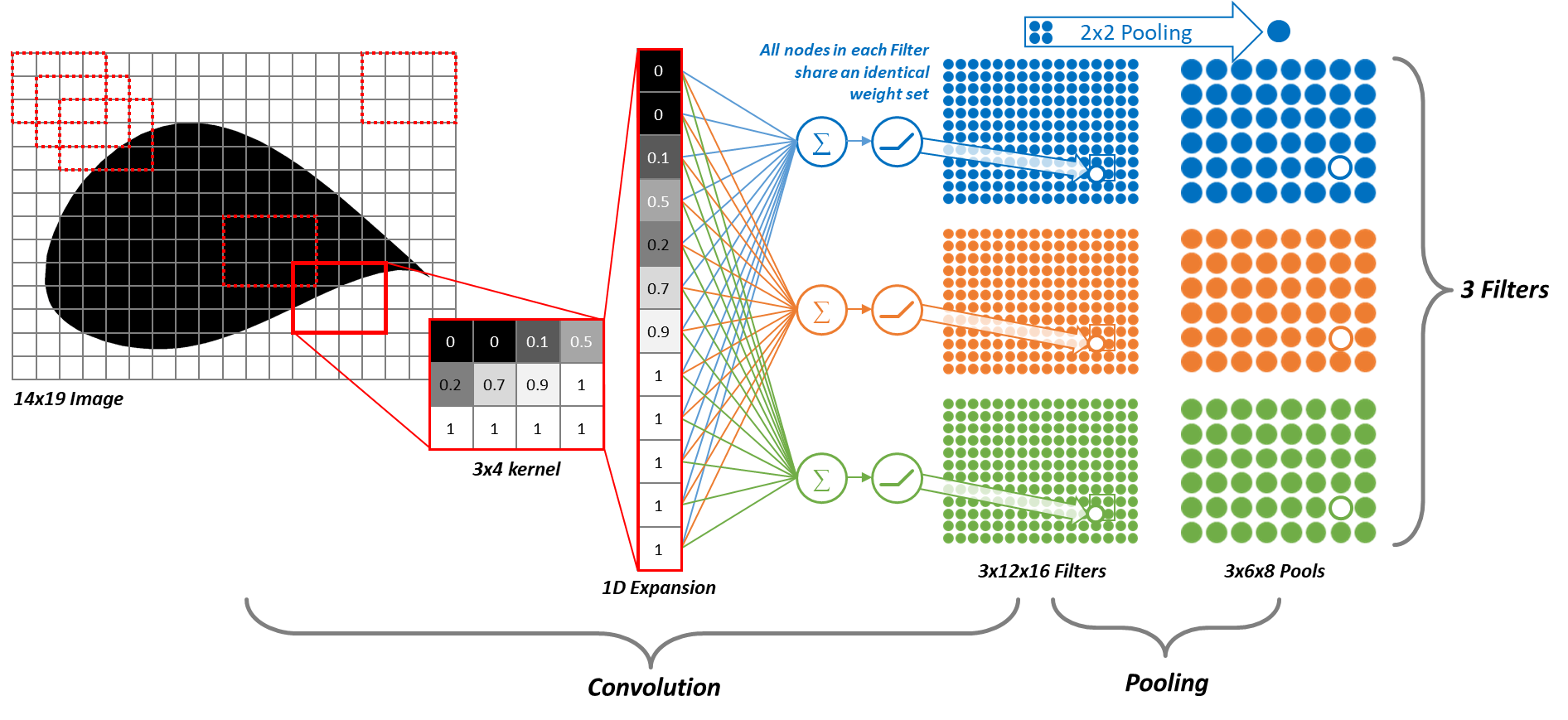}
\caption{Illustration of convolution scheme (Numbers are not actual, but only for illustration purpose.)}
\label{fig:conv}
\end{figure}

For the actual implementation, three different neural network structures were designed to learn and predict the lift coefficient of the airfoil, as shown in Fig. \ref{fig:3nets}. The `MLP' represents a conventional multi-layered perceptron, which has two hidden layers between the input and output layers. Here `FC' means a fully connected layer where each and every units in one layer connects to each and every units in the other layer. In this work, the \emph{MLP} serves as a baseline architecture to assess the relative performance of the other networks. `AeroCNN-I' and `AeroCNN-II' are two types of simple CNN architectures with different convolution schemes and internal layouts. Each architecture is designed to process a unique input data structure, hence, requiring different process for its data preparation. \emph{MLP} uses the freestream Mach number, Reynolds number, angle of attack, and $100$ $y$-coordinates ($50$ points for upper surface and $50$ points for lower surface) around each airfoil at pre-defined $50$ $x$-positions. These 50 positions are mildly clustered toward the leading edge considering the higher variability in airfoil geometry near its leading edge. This results in total $103$ input parameters. \emph{AeroCNN-I} has the same 3 parameters for the flow condition and the same $100$ $y$-coordinates. However, the $y$-coordinates has a form of the 2D array $(2 \times 50)$ structure convoluted by the local 2D kernels having the dimension of $(2 \times 5)$. Here, the first row corresponds to the lower surface $y$-coordinates and the second row to the upper surface. \emph{AeroCNN-I} is an intermediate architecture between the conventional MLP and fully-convoluting \emph{AeroCNN-II}. Fig. \ref{fig:conv} illustrates a typical convolution scheme used in image recognition tasks where the local receptive field with a kernel of $(3 \times 4)$ maps to $3$ filters that are sub-sampled by $(2 \times 2)$ pooling. This `pooling' operation is usually executed, in order to reduce dimensionality of the learning process, by averaging or taking the maximum value. For \emph{AeroCNN-II} only, an `artificial image' for each data entry is pre-processed to $(49 \times 49)$ pixels of 2D array. $(5 \times 5)$ kernels, $25$ filters, and $(2 \times 2)$ `max' pooling have been chosen after a series of numerical experiments to ensure a reasonable prediction performance. The number of `convolution-pooling' pairs has also been chosen as one by the same reason, i.e., two pairs of `convolution-pooling' that is typical in many image recognition tasks doesn't provide a significant benefit in this case.

\begin{figure}[h!]
\centering
\includegraphics[width=0.95\textwidth]{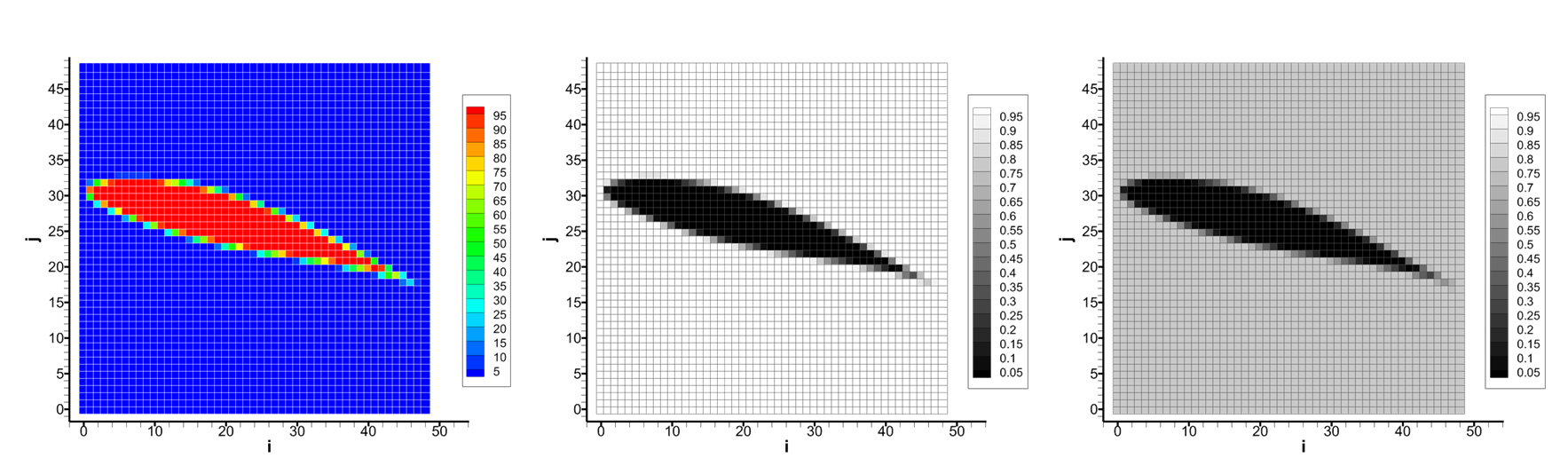}
\caption{`Artificial image' of NACA654421a05 at $24^o$-AoA; (left) tilt by AoA and digitization by pixel density, (middle) nondimensionalized for Mach 0.8, and (right) the same geometry for Mach 0.6}
\label{fig:image}
\end{figure}

\emph{AeroCNN-II} uses a full image-like 2D array combining the airfoil shape and the flow conditions. For \emph{AeroCNN-II}, to automate the data preparation process and infuse the flow conditions in unified manner, instead of collecting or rasterizing an `image' of an airfoil, a unique concept of `artificial image' has been devised and applied to each data entry. The airfoil shape is tilted for the corresponding angle of attack, after which the airfoil shape is converted to an image-like 2D array having a finite number of rows and columns, $(49 \times 49)$ in this work. Finally, the external space is `colored' by following equation depending on the freestream Mach number and the pixel density;
\begin{equation}
\bar{\rho} = \left(1-\frac{\rho}{\rho_{max}}\right)\frac{M_\infty}{M_{\infty,max}}
\end{equation}
where the raw pixel density, $\rho$, for each cell has the range $[0,100]$; $100$ for the cell located `completely' inside an airfoil, $0$ for the cell located `completely' outside an airfoil, and $(0,100)$ for the cell located on the boundary of an airfoil shape. By this mapping, for each airfoil shape, the final pixel density, $\bar{\rho}$, has the common minimum value of $0$ for the cell located `completely' inside the airfoil, but the higher maximum value for the higher freestream Mach number. The overall process is illustrated in Fig. \ref{fig:image} where the external points have different `color' depending on the corresponding freestream Mach numbers. In this way, the angle of attack, the freestream Mach number, and the airfoil shape are all combined into single data format of the `artificial image.' For \emph{AeroCNN-II}, the Reynolds number is excluded and only one subset of the entire dataset has been used, i.e., for a single Reynolds number. 

All three architectures have been implemented in the Python environment using a modern, open-source library for deep learning: MXNET (http://mxnet.incubator.apache.org).

\subsection{Data Preparation}
\begin{figure}[h!]
\centering
\includegraphics[width=0.7\textwidth]{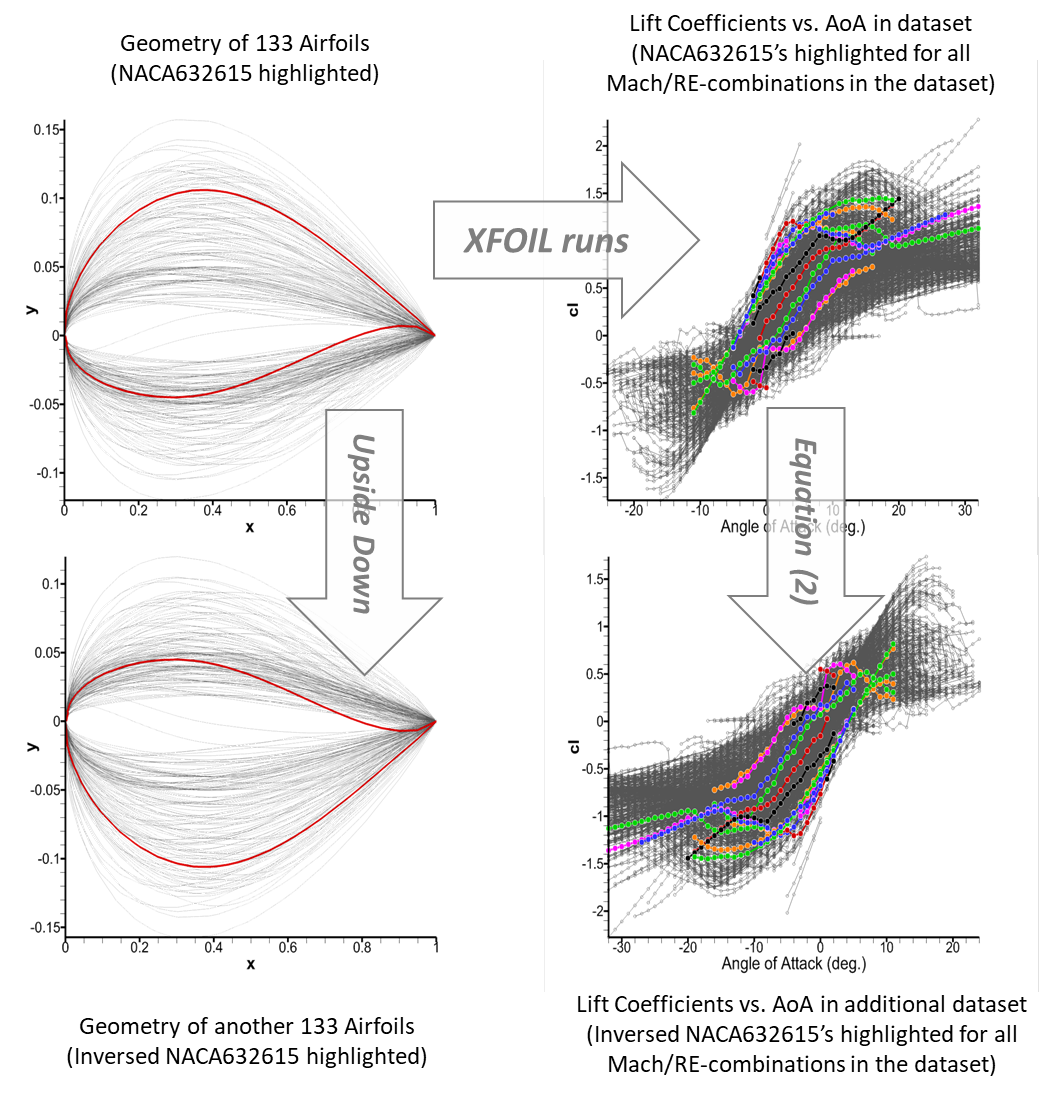}
\caption{Data preparation process}
\label{fig:data}
\end{figure}

The overall data preparation procedures are illustrated in Fig. \ref{fig:data}. Obtaining an accurate airfoil shape is critical for the network training results. There are different ways to represent the airfoil shape. The most neutral and common way is to represent it as coordinate points along the upper and the lower surfaces. In this work, the UIUC Airfoil Data Site \cite{selig1996uiuc} serves as the dataset necessary to obtain the airfoil shapes. The database has a collection of point-by-point airfoil coordinates format, with $(x,y)$ coordinates. A total $133$ sets of 2D airfoil geometry are used as training and validation datasets. This includes a symmetric airfoil, NACA 4-digit and 6-digit series, and others. All airfoils have outer model line shapes without any high lifting apparatus.

In this work, the aerodynamic coefficient of interest is focused on the lift force coefficient, $C_l$, to prove the concept of the technique. The collection of the aerodynamic data have been generated from a series of numerical simulations. Due to the nature of this work, the absolute accuracy of the simulation results is not the first priority. Considering the required diversity in flow conditions and airfoil shapes, XFOIL \cite{drela1989xfoil} was chosen as the simulation tool to obtain the aerodynamic coefficients. The lift coefficients were obtained for diverse flow conditions for each airfoil using XFOIL as a function of angle of attack $(-10^o - 30^o)$, Reynolds number (30,000 - 6,500,000), and freestream Mach number $(0.3 - 0.8)$. By running XFOIL and eliminating diverging or non-converging cases, the total number of data entries is about 40,000. To augment the data quality as well as quantity, for each airfoil with a given combination of Mach and Reynolds number, its `upside-down' or flipped airfoil data has also been created using symmetry in the lift force;
\begin{equation}
C_{l,\text{upside-down}}(\alpha)=-C_{l,\text{original}}(-\alpha)
\end{equation}
This operation makes the size of entire dataset increase to about 80,000.

\subsection{Activation Functions}
\begin{figure}[h!]
\centering
\includegraphics[width=0.5\textwidth]{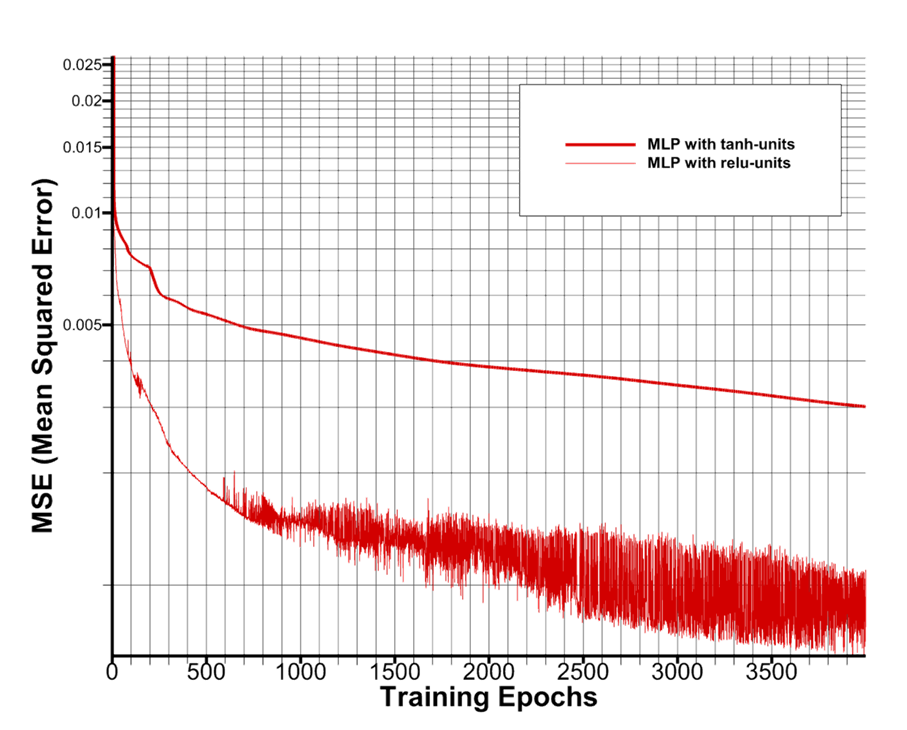}
\caption{Typical \emph{MLP} training history with \emph{tanh} and \emph{ReLU} activation}
\label{fig:relu}
\end{figure}

In a traditional MLP approach, the hyperbolic tangent (tanh) prevails as the most common choice for the activation function of hidden units. Elman \cite{elman1993learning} pointed out that the functional shape of the sigmoid-type activators is blamable for the later stagnation in the training process due to its vanishing (or infinitesimal) gradients at the two extrema. Krizhevsky et al. \cite{krizhevsky2012imagenet} showed the benefit of replacing \emph{tanh} units by the rectified linear unit (ReLU) in image recognition tasks. In the present work, a variety of network structure was tested with both \emph{tanh} and \emph{ReLU} units as the main hidden unit. One training history of the MLP network is shown as a typical example in Fig. \ref{fig:relu}. Here, the MLP network has $103$ input units corresponding to the parameters for airfoil geometry and flow conditions, two hidden layers with $200$ and $100$ hidden units, and a single output unit of the lift coefficient. It can be seen that the learning capability is significantly higher with the \emph{ReLU} unit although there is an oscillatory behavior. The almost identical benefit with Krizhevsky et al. \cite{krizhevsky2012imagenet} is obvious for the current task of learning lift coefficients. Therefore, all \emph{MLP}, \emph{AeroCNN-I} and \emph{AeroCNN-II} networks adopt the rectified linear unit as the only activation function except for the pooling and linear units.

\section{Results and Discussion}
\begin{figure}[h!]
\centering
\begin{subfigure}{.5\textwidth}
\centering
\includegraphics[width=0.95\linewidth]{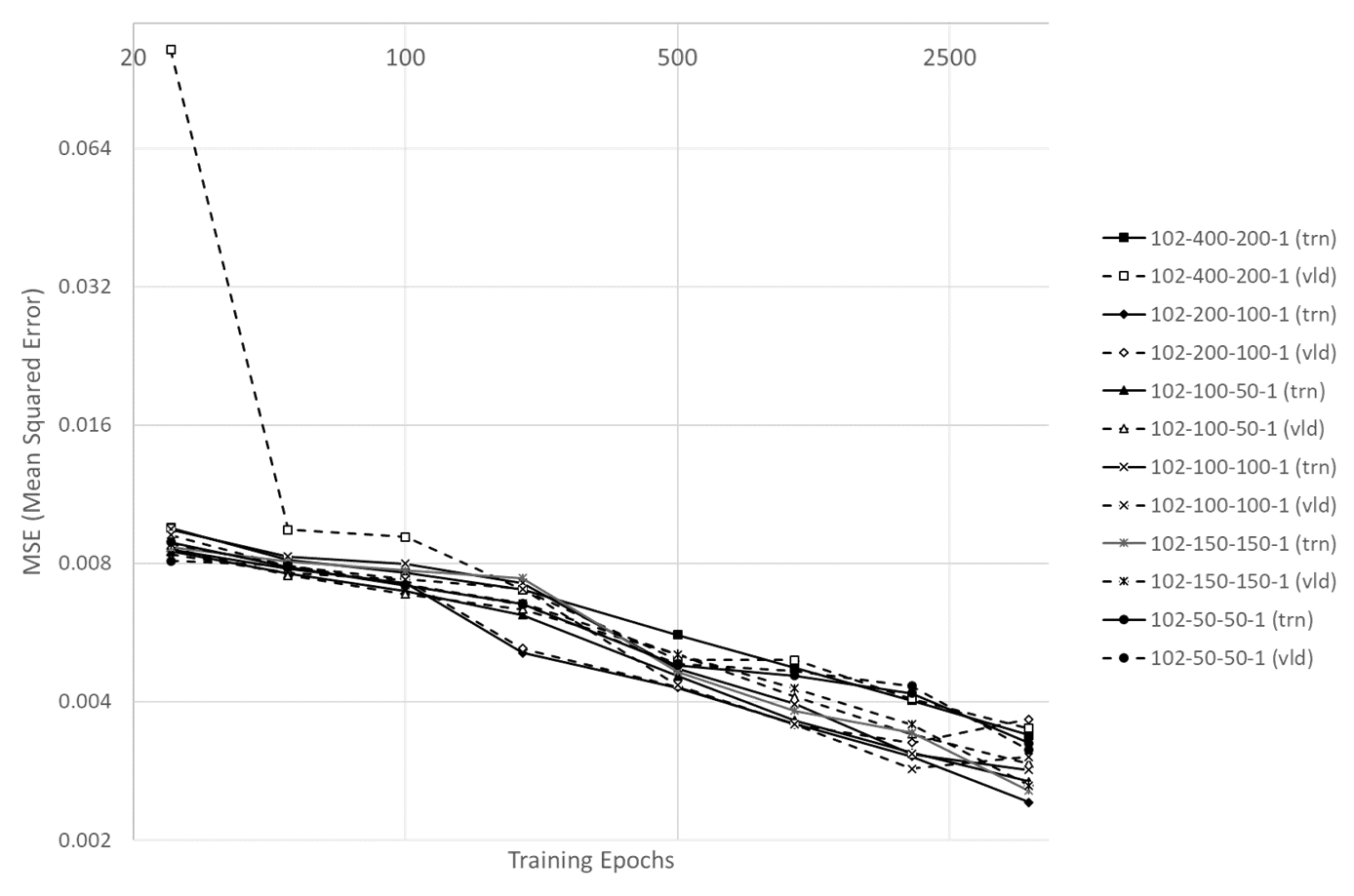}
\caption{various \emph{MLP} networks}
\label{fig:mlps}
\end{subfigure}%
\begin{subfigure}{.5\textwidth}
\centering
\includegraphics[width=0.95\linewidth]{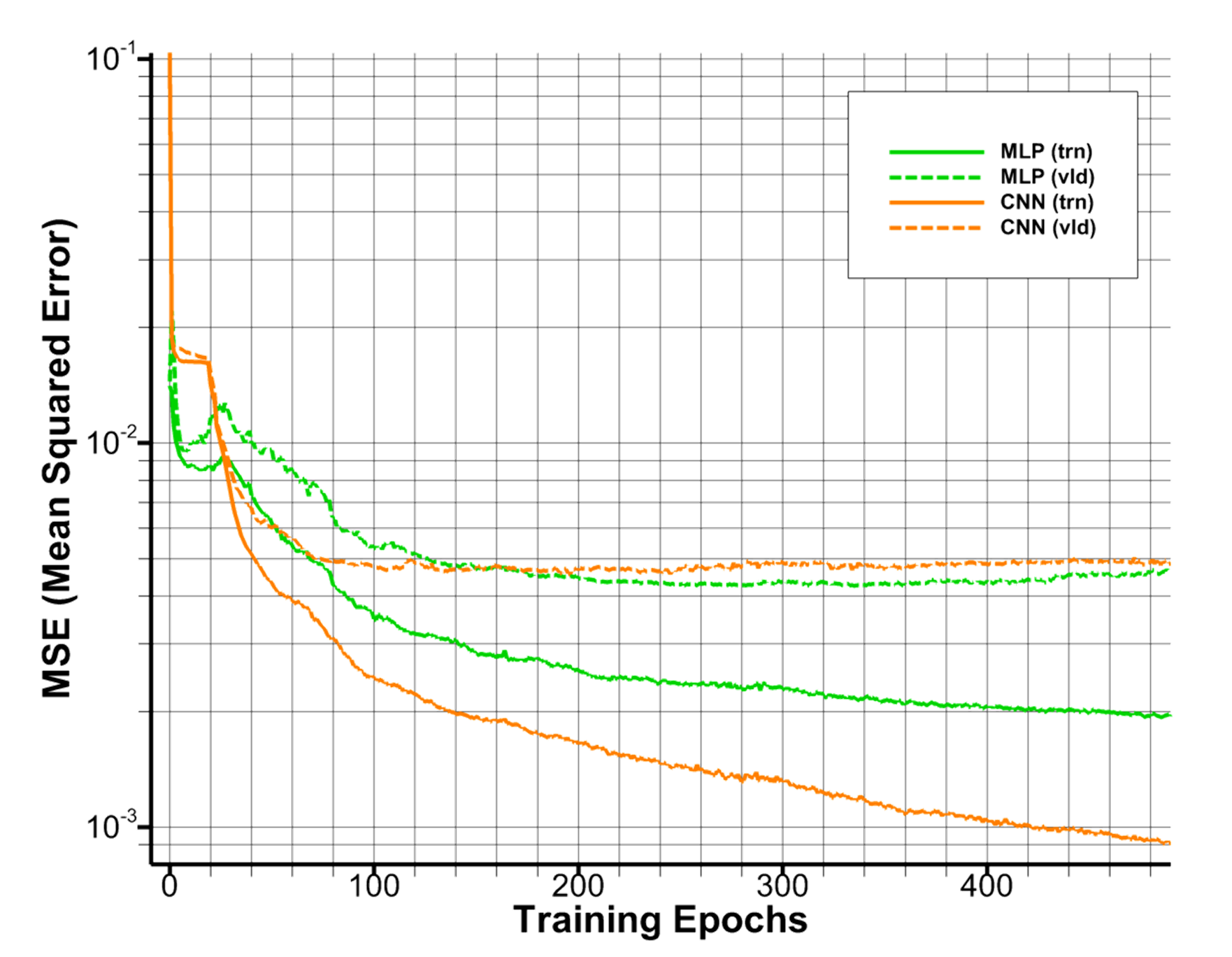}
\caption{\emph{MLP} and \emph{AeroCNN-I}}
\label{fig:cnn_I}
\end{subfigure}
\caption{Training history of \emph{MLP} and \emph{AeroCNN-I}}
\end{figure}

Both for the MLP and CNNs, weight and biases in the neural networks were trained by \emph{AdaDelta} algorithm \cite{zeiler2012adadelta} that is built in MXNET library, which is a popular choice for modern neural network training due to its flexibility and robustness. It is known that \emph{AdaDelta} algorithm has shown superb error-reduction capability with notable insensitivity to the choices in the hyper-parameters for weight optimization \cite{zeiler2012adadelta}. Under MLP architecture, various structures with different numbers of hidden units have been tested to obtain the overall boundary of the training performance. Fig. \ref{fig:mlps} presents the training history of those models with a converged mean squared error (MSE) for the training $(85\%)$ and validation $(15\%)$ datasets. Regardless of the diversity in the number of adjustable parameters or learning capabilities, they show similar learning trajectories resulting in the overall boundary of the obtainable accuracy for the lift coefficient prediction.

Fig. \ref{fig:cnn_I} compares the learning capability between \emph{MLP} and \emph{AeroCNN-I} architectures. The two models have approximately same number of adjustable parameters. In this case, the $30\%$ and $70\%$ split of the entire dataset was used for the training and validation datasets, respectively. It can be seen that, when comparing \emph{MLP} with \emph{AeroCNN-I}, \emph{AeroCNN-I} curves drop faster for both training and validation datasets, before reaching steady MSEs. This implies that \emph{AeroCNN-I} learns faster than \emph{MLP} for the given number of training epochs. Here, an `epoch' means one full exposure of whole training examples to the learning architecture. However, the drawback of \emph{AeroCNN-I} architecture is the time consumption. In this case, the total time spent for $500$ epochs was 3,206 seconds for \emph{AeroCNN-I} while only 1,700 seconds were necessary for \emph{MLP}. The CNN architecture requires significantly more calculation steps and time per epoch than the MLP architecture for the similar number of adjustable parameters. 

\begin{figure}[h!]
\centering
\begin{subfigure}{.5\textwidth}
\centering
\includegraphics[width=0.99\linewidth]{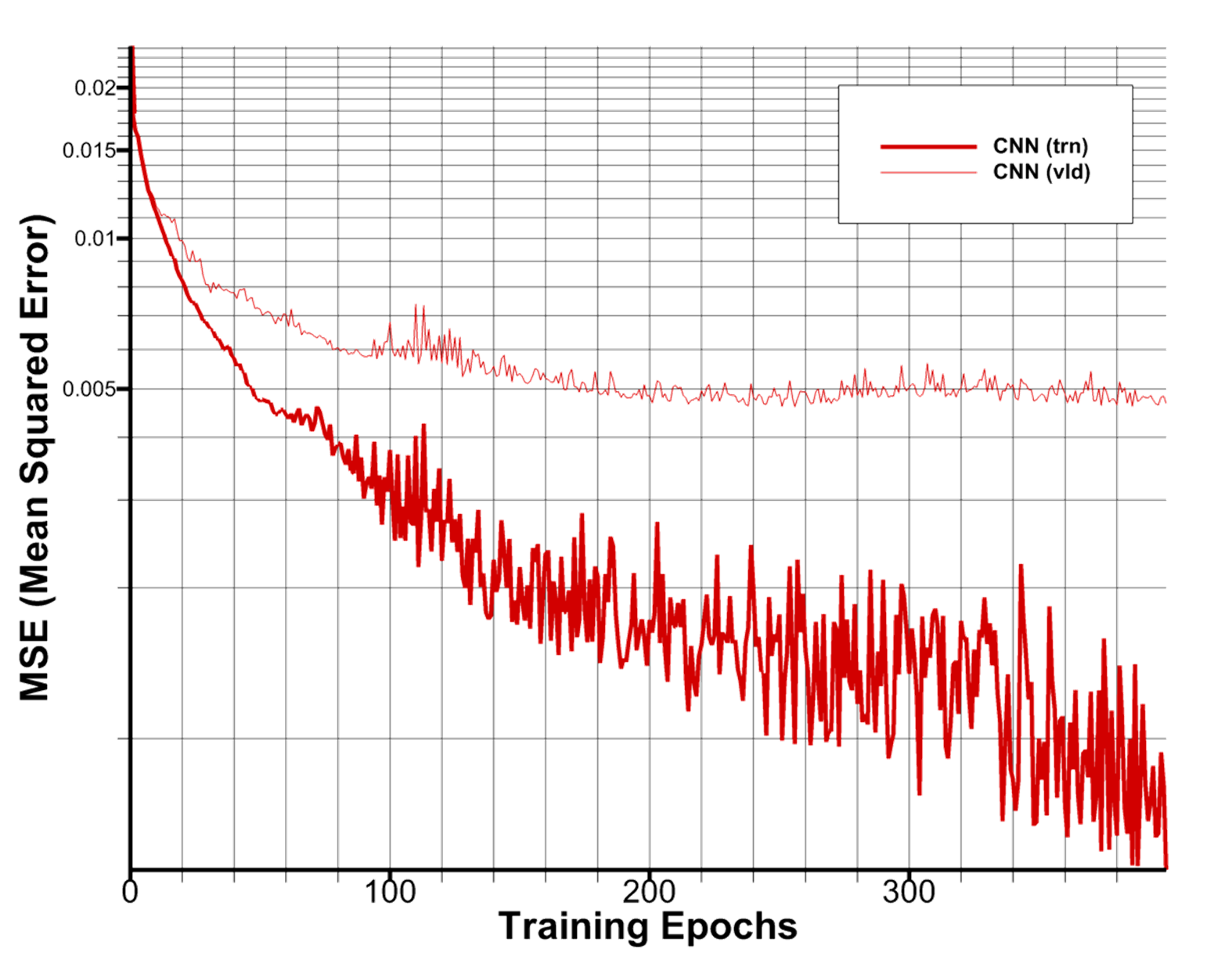}
\caption{Training history}
\label{fig:cnn_II}
\end{subfigure}%
\begin{subfigure}{.5\textwidth}
\centering
\includegraphics[width=0.99\linewidth]{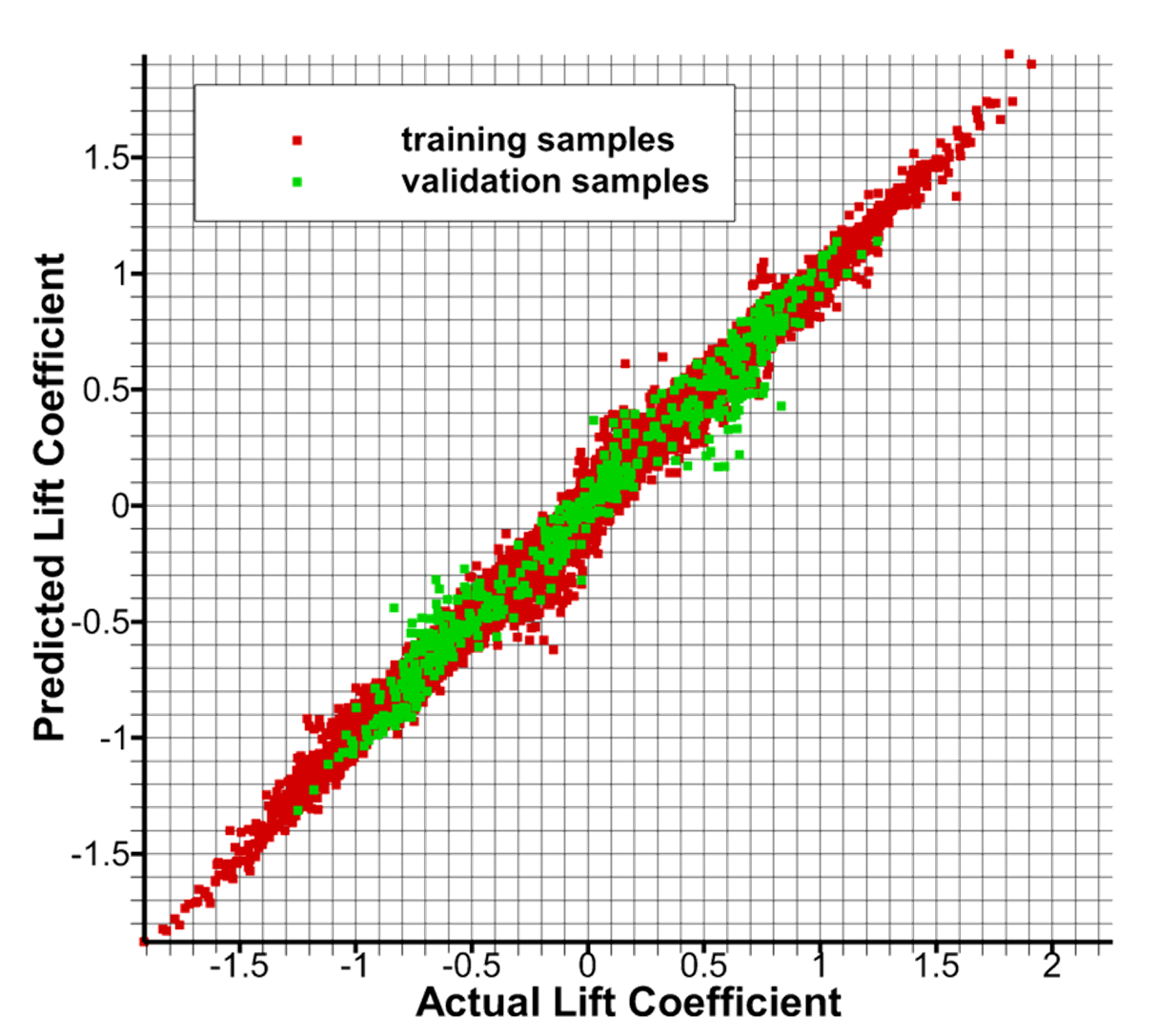}
\caption{Actual versus predicted lift coefficients}
\label{fig:cl}
\end{subfigure}
\caption{Results of \emph{AeroCNN-II} training}
\end{figure}

The training and validation results from \emph{AeroCNN-II} architecture are plotted in Fig. \ref{fig:cnn_II}. \emph{AeroCNN-II} uses full 2D-convolution for the pre-processed `artificial images.' From Fig. \ref{fig:cnn_II}, the performance from \emph{AeroCNN-II} shows a similar training trajectories and comparable results as does \emph{AeroCNN-I} in Fig. \ref{fig:cnn_I}. This result suggests that the proposed architecture of \emph{AeroCNN-II} by incorporating the flow conditions and airfoil geometry into the `artificial image' is a working concept. 

\begin{figure}[h!]
\centering
\includegraphics[width=0.55\textwidth]{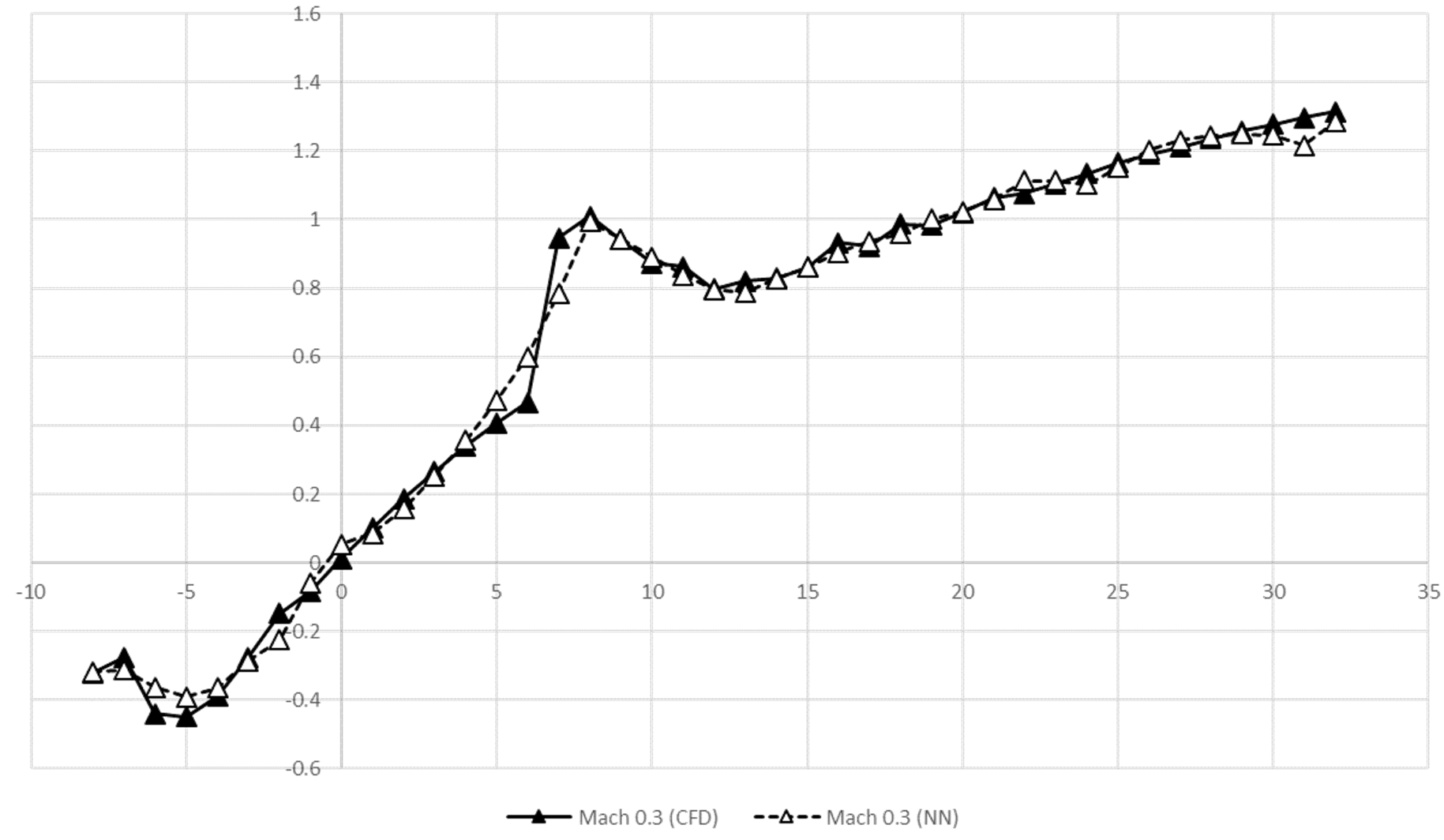}
\caption{Actual versus predicted $C_l-\alpha$ curve at $M_\infty=0.3$ for NACA64a410 (included in training dataset)}
\label{fig:trn1}
\end{figure}

\begin{figure}[h!]
\centering
\begin{subfigure}{.5\textwidth}
\centering
\includegraphics[width=0.99\linewidth]{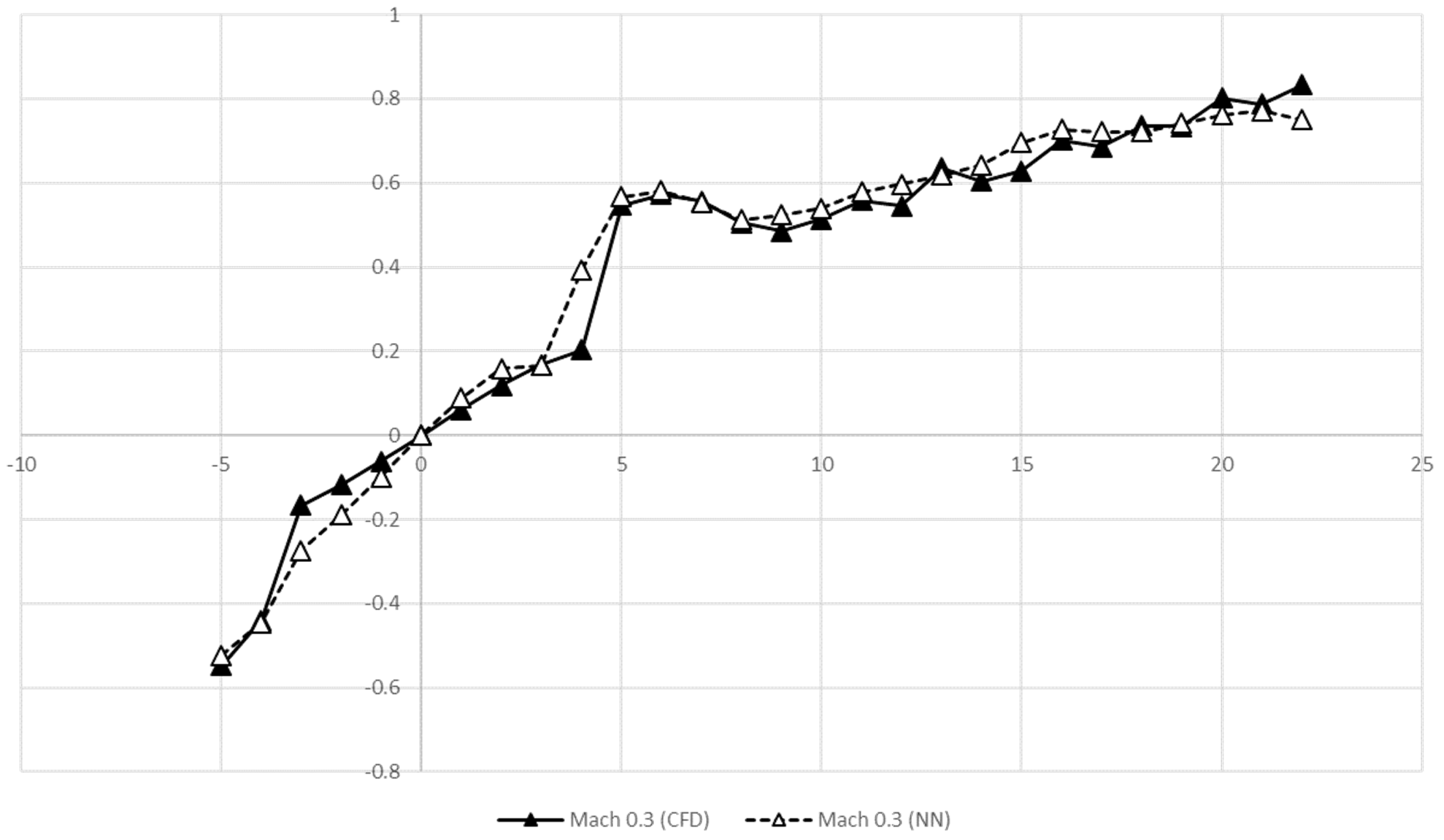}
\caption{Geo444 $(M_\infty=0.3)$}
\label{fig:vld1}
\end{subfigure}%
\begin{subfigure}{.5\textwidth}
\centering
\includegraphics[width=0.99\linewidth]{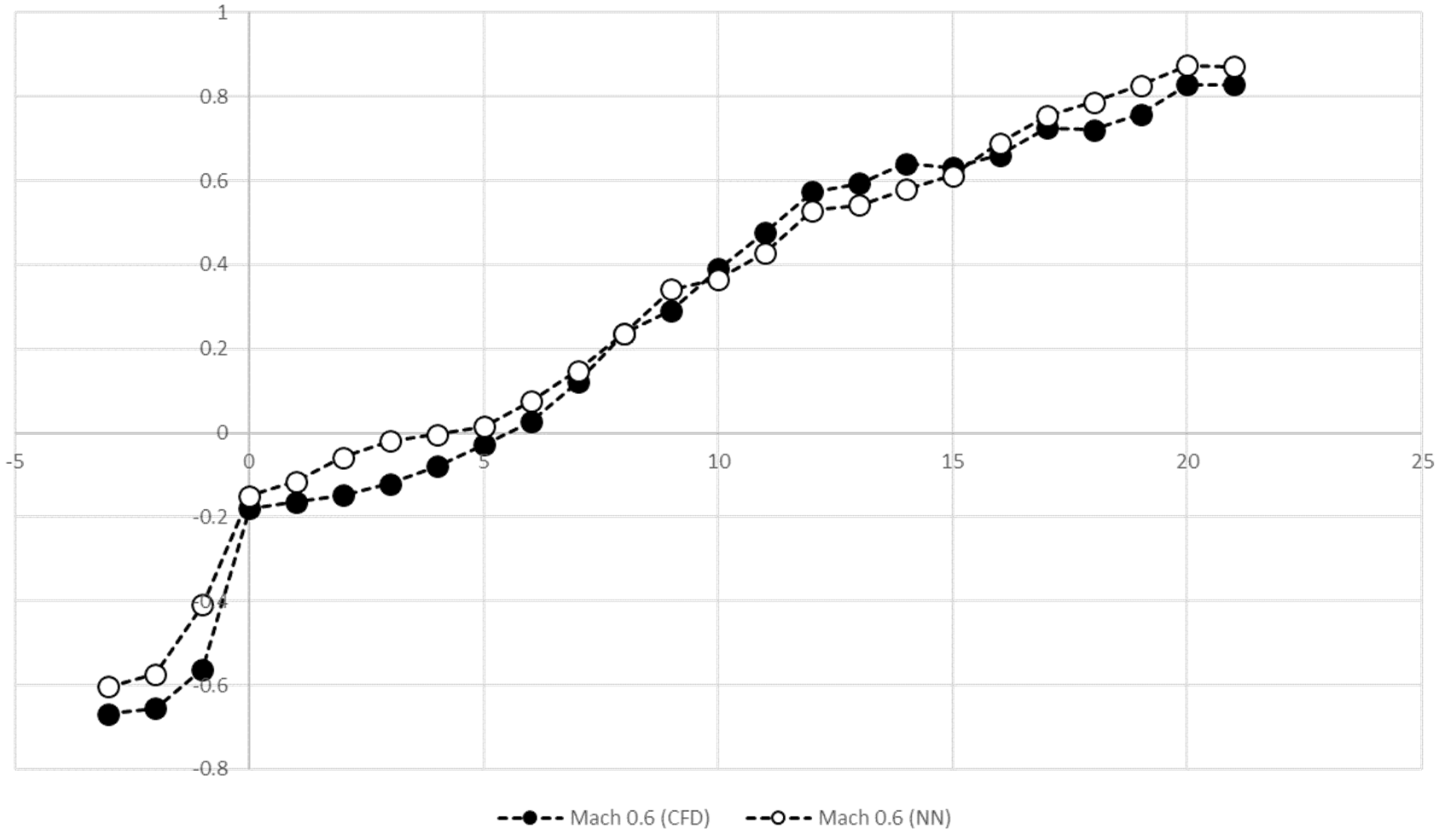}
\caption{NACA65415a05 $(M_\infty=0.6)$}
\label{fig:vld2}
\end{subfigure}
\caption{Actual versus predicted $C_l-\alpha$ curve (not included in training dataset)}
\end{figure}

A reduced subset of entire dataset for a single Reynolds number was processed through \emph{AeroCNN-II} with $95\%$ of the data used for training and $5\%$ used for validation. The results of the predicted lift coefficients were plotted against the actual lift coefficients obtained from the XFOIL runs. As shown in Fig. \ref{fig:cl}, the majority of points are clustered near the $45^o$-line, meaning that the predicted values are close to the actual values. The training samples indicated by red color fit better than the validation samples indicated by green color. This is a common result in many adaptive machine learning techniques; the generalization for unseen examples is more difficult than just reducing training error. In order to test the prediction capability of \emph{AeroCNN-II}, the lift coefficients are plotted as a function  of $\alpha$ for both the CFD results and the prediction from \emph{AeroCNN-II}. Fig. \ref{fig:trn1} is the $c_l-\alpha$ curve for an airfoil that is included in the training dataset. Fig. \ref{fig:vld1} and Fig. \ref{fig:vld2} are for unseen airfoils during the training process. These plots exhibit reasonable training and generalization performance of \emph{AeroCNN-II} architecture for the given task of predicting lift coefficients.

In summary, as a proof of concept, the proposed CNN architectures demonstrates reasonable performance in learning from the given examples and in predicting lift coefficients for unseen airfoil shapes.

\section{Conclusion}
Various CNN architecture were applied to study aerodynamics. In this work, they were being applied to predict the lift coefficient with a given airfoil shape and flow conditions. Predictions from both the conventional MLP and the proposed CNN techniques were generated and compared. Two types of CNN architectures were constructed with different convolution schemes and internal layouts. \emph{AeroCNN-II} is the first of its kind to investigate the 2D aerodynamic problems involving diverse flow conditions and the variety of sectional shapes in a single framework of the CNN using the concept of the `artificial image.' 

The synthesis between geometric boundary conditions (e.g., airfoil shape) and non-geometric boundary conditions (e.g., $\alpha$ and $M_\infty$) into an image-like array has been successfully used in training the CNN architecture for aerodynamic meta-modeling task. This result gives a useful perspective to harness well-developed deep learning techniques in image recognition tasks for engineering meta-modeling tasks.

\newpage

\bibliography{sample}

\end{document}